# Using Orthophoto for Building Boundary Sharpening in the Digital Surface Model

**Xiaohu Lu** [a], PhD Student
lu.2037@osu.edu
**Rongjun Qin** [a,b], Professor
qin.324@osu.edu
**Xu Huang** [a], Post Dr.
huangxu.chess@163.com
[a] Department of Civil, Environmental and Geodetic Engineering, The Ohio State University,
218B Bolz Hall, 2036 Neil Avenue, Columbus, OH 43210, USA;
[b] Department of Electrical and Computer Engineering, The Ohio State University,
205 Dreese Labs, 2015 Neil Avenue, Columbus, OH 43210, USA

## ABSTRACT

Nowadays dense stereo matching has become one of the dominant tools in 3D reconstruction of urban regions for its low cost and high flexibility in generating dense 3D points. However, state-of-the-art stereo matching algorithms usually apply a semi-global matching (SGM) strategy. This strategy normally assumes the surface geometry pieces-wise planar (Hirschmuller, 2008), where a smooth penalty is imposed to deal with non-texture or repeating-texture areas. This on one hand, generates much smooth surface models, while on the other hand, may partially leads to smoothing on depth discontinuities, particularly for fence-shaped regions or densely built areas with narrow streets. To solve this problem, in this work, we propose to use the line segment information extracted from the corresponding orthophoto as a pose-processing tool to sharpen the building boundary of the Digital Surface Model (DSM) generated by SGM. Two methods which are based on graph-cut and plane fitting are proposed and compared. Experimental results on several satellite datasets with ground truth show the robustness and effectiveness of the proposed DSM sharpening method.

## INTRODUCTION

The increasing availability and resolution of the space-borne imagery are drawing great attention in generating valuable 3D content of the terrain and ground object. Typically, stereo matching algorithms are applied to generate DSM for those 3D content, which usually smooths the depth discontinuities on building areas with narrow streets. Although ad-hoc or adaptive tuning of the smooth penalty parameters may seem to work to a certain degree, while it is often a research question of how these parameters should be appropriately selected. Given that in a top-view mapping scenario, the orthophoto corrected by the produced digital surface model often preserve building boundaries well, it is reasonable to consider a post-process strategy that directly leverage the consistency on the building boundaries both on the DSM and orthophoto.

Related post-processing methods generally take pixel and contour/line as primitives (pixel-based and contour line based). The pixel based methods (Huang and Zhang, 2016; Park et al., 2015) normally give a support window centered at pixels on the boundaries which defines the neighborhood of each pixel, and utilize the disparity of neighbor intensity-similar pixels to refine the disparity of central pixel. For example, in the work of (Huang and Zhang, 2016), the possibility of correct disparity value in the neighborhood is calculated and aggregated first to generate a belief value first, then those belief values are propagated such that pixel having lower aggregated belief will receive propagation from the higher belief pixel with similar color. The contour/line based methods (Qin et al., 2018; Hirschmuller, 2008; Li et al., 2015) refine the entire boundary set instead of individual pixels. They usually extract building boundaries or line segments first, and then use plane function to refine the disparity of these boundaries. For example, in the work of (Qin et al., 2018) line segments are firstly extracted and matched for the two input images, then planes are fitted based on disparity for those matched lines, finally a plane-based adjustment procedure is applied to enhance the disparity. There are also some researches on using external information to refine the DSM boundary, for example the building footprints are utilized in the work of (Brédif et al. 2013) to sharpen the DSM building boundary. However, those methods might be limited by the availability of the data and the registration accuracy.

Straight line segments (Lu et al., 2015; Von Gioi et al. 2010) on the orthophoto preserves important information about the building boundary, which can be used to cut out the falsely smoothed building boundary pixels and make sure that straight instead of zigzag boundary can be obtained in the DSM. Therefore, in this work, we proposed two methods that takes line segments extracted from orthophoto as constrains to adjust the DSM such that erroneous depth values on the boundary will be corrected and shape of boundary will be clearer. The rest of this paper is organized as follows: Section 2 introduces the proposed methodology in detail, which is followed by the experimental results in Section 3, and finally the conclusions are drawn in Section 4.

## METHODOLOGY

In general, the proposed method takes a DSM and its corresponding orthophoto as input, then the building boundaries and line segments are extracted from the DSM and orthophoto images, respectively. After that, one method which utilizes the graph cut algorithm to adjust the position of the DSM boundary pixels, and another method which applies plane fitting to adjust the values of DSM boundary pixels are proposed to sharpen the building boundary.

**Pre-Processing**

To refine the building boundary, firstly we need to identify the boundary in the DSM and orthophoto. However, general line segment detector like LSD (Von Gioi et al. 2010) and CannyLines (Lu et al., 2015) will not only extract the line segments on the boundary but also those on the roof or the ground. To eliminate those non-building line segments, we propose a solution to use the building boundary extracted from the DSM to filter out those outlier line segments, which contains the following two steps:

(1) DSM Building Boundary Extraction

For DSM building boundary extraction, we applied the morphological tophat operation on the DSM, which will find out the bright spots of the image that are smaller than the structuring element. However, since the sizes of objects on the DSM vary a lot, a single constant structuring element will probably not be able to extract all the objects, we thus adopt a robust tophat operation which combines the outputs of a series of structuring element together to form the final result. In our experiment, the sizes of structuring element range [10, 400] with an interval of 10 pixels. With the final result of tophat operation, which is a binary mask of building segments, the building boundaries can be easily extracted by contour detection. Figure 1(b) shows the final result of tophat operation on one of the resting regions.

(2) Orthophoto Building Line Extraction

For building line extraction on the orthophoto, we utilize the LSD detector which can generate reasonable line segments as well as eliminate the outliers. To filter out those non-building line segments, we utilize the building boundaries extracted from the above step, and then create a buffer zone for those boundaries to identify the building boundary region, which results a binary image with 1 denotes building boundary region, 0 denotes other. Line segments with more than 50 percentages lay with the buffer region will be kept as the building line segments, while the others are removed as non-building ones. Figure 1(c), (d) shows the original and filtered line segments on one of the testing regions. As can be seen from it that the DSM boundary buffer zone is a good information to get rid of those non-building line segments.

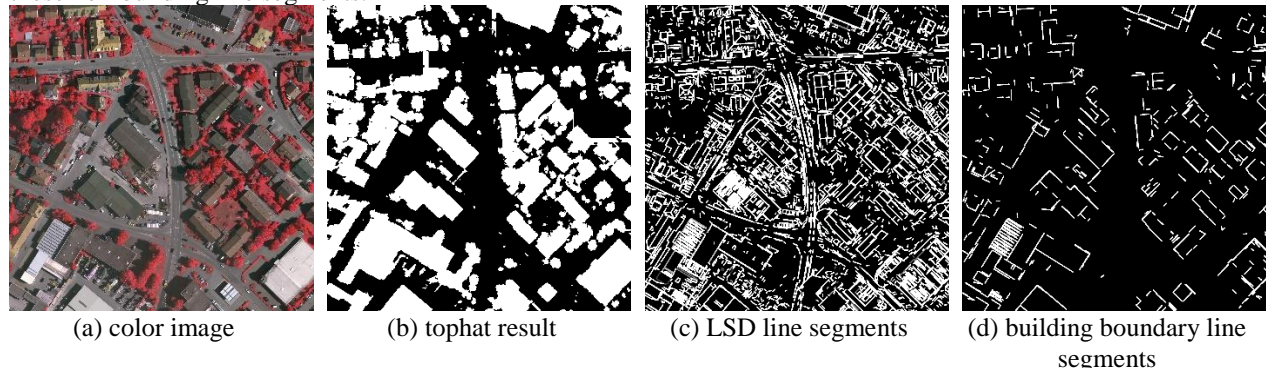

(a) color image     (b) tophat result     (c) LSD line segments     (d) building boundary line segments

Figure 1 Pre-processing results on the testing region. The amount of original line segments on (c) is 4918, while that of the filtered line segments on (d) is 610.

**Boundary Adjustment**

Given the filtered line segments, we want to adjust the DSM pixels to make the boundary as sharp as possible. To achieve this goal, one can shift the position of DSM boundary pixels to make them align with the straight line, one can also modify the value of DSM boundary pixels on each side of the line segment to enforce there to be a line segment on the DSM. For the first one, we propose a graph-cut based adjustment method. While for the second one , a line based adjustment is proposed.

**(1) Graph-cut Based Boundary Adjustment**

Graph-cut is (Boykov et al., 2001) can be used to solve the energy function minimizing problem as follows:

$$E(l) = \sum_p D(p, l_p) + \sum_{p,q} V_{pq}(l_p, l_q),$$

Here we have a finite set of data points **P** and a finite set of labels **L**, $D(p, l_p)$ is the cost of assigning data point p with a label $l_p$, which is known as the data term, $V_{pq}(l_p, l_q)$ is the cost of assigning neighboring pixels (p, q) with labels $(l_p, l_q)$, which is known as the smooth term. The graph-cut then utilizes the max flow algorithm to find out the optimal solution which minimize the energy function. In this case, data points **P** is the contour pixels extracted by tophat on the DSM, the label set **L** is defined as the offset of contour pixels, which is a $11 \times 11$ grid centered on the pixel as follows:

| (-5, -5) | (-4, -5) | ... | (4, -5) | (5, -5) |
|---|---|---|---|---|
| (-5, -4) | (-4, -4) | ... | (4, -4) | (5, -4) |
| ... | | (0, 0) | | |
| (-5, 4) | (-4, 4) | ... | (4, 4) | (5, 4) |
| (-5, 5) | (-4, 5) | ... | (4, 5) | (5, 5) |

Figure 2 The label set **L** is defined as a $11 \times 11$ 2d offset grid centered on the pixel.

For each data point $p = (x_p, y_p)$, the cost of assigning it label $l_p = (x_l, y_l)$ can be defined as follows:

$$D(p, l_p) = \begin{cases} 0, & if \ (x_p + x_l, y_p + y_l) \ is \ in \ line \ buffer \\ 10, & otherwise \end{cases}$$

Here the line buffer is generated based on the filtered building line segments with a buffer size of 2 pixels. To make sure that pixels on the same contour are continuous after the optimization, we also add the smooth term which is defined as follows:

$$V_{pq}(l_p, l_q) = \begin{cases} 2, if \ \| l_p - l_q \| < 5.0 \\ 100, otherwise \end{cases}$$

Here q is the pixels in data point p's neighborhood which is defined as the following 8 pixels of data point p in the contour sequence. The smooth term is designed to make sure that the neighboring data points on the same contour should not have very large difference on their label, thus the resulting contour after optimization will not be broken. Having defined the data term and the smooth term, the graph-cut can then solve the energy function minimization efficiently within a few seconds. After that, for each boundary pixel, we can get its offset vector from its label, we also set the offset vector of those internal pixels far away (20 pixels) from the boundaries as (0,0) because they are not supposed to be adjusted. The offset vectors of all the remaining pixels are thus interpolated by the offset vectors of those boundary and internal pixels. Finally, the adjusted DSM can be formed by shifting the pixels on the original DSM with their corresponding offset values. Figure 3 demonstrates the result of graph-cut based boundary adjustment method on a testing image. As can be seen that the graph-cut method can keeps the well-aligned boundary pixels unchanged, while adjust other pixels well with the image.

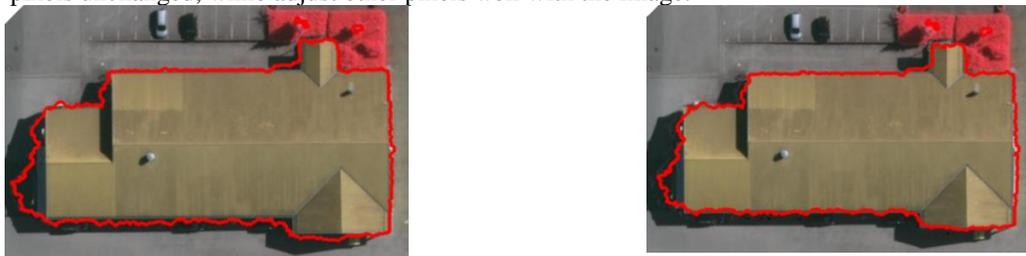

(a) original boundary          (b) graph-cut boundary

Figure 3 Comparison between the DSM boundary before and after graph-cut based adjustment. Boundary pixels on the bottom are well aligned with the image after optimization.

**(2) Line Based Boundary Adjustment**

To further explore methods that can be useful for DSM boundary sharpening, we also proposed another line based boundary adjustment method which directly utilizes the line segments extracted form the orthophoto to guide the fitting of plane on the DSM.

More specifically, given a line segment extracted from the orthophoto, we firstly use a rectangle buffer to define its neighborhood. Only pixels inside the neighborhood will be considered for plane fitting. Considering the fact that the distance between buildings varies a lot in different region, it is reasonable to set different values of buffer size for different line segments to get better result. We thus proposed an adaptive strategy for the determination of the buffer size. As introduced in the pre-processing procedure, line segments are filtered by the DSM boundaries. Those boundaries are generated via tophat with different sizes of structuring element, which means that we can somehow utilize the size of the structuring element to represent the width of the building. The implementation details are as follows. Firstly, for each value of structuring element size ranging from [10, 400] with an interval of 10 pixels, we combine its result with all the pervious results to generate the binary mask image and then extract contours from the combined mask and save the contour to an image. Then, given each line segment we traverse all those contour images to see that on which contour image does the line segment is within the buffer area of the contours for the first time, and set the width of the line segment as the index of the contour image. That is to say, if a line segment is overlapped with contours on the first contour image, we will set the line's width as 1, while that will be 40 for the last contour image. Finally, the buffer size is set as $min(3 \times line\ width,\ 30)$ to avoid too wide buffer area.

Having defined the buffer area of each line segment, we then collect pixels from the DSM image on each side of the line segment in the buffer area to fit for plane. The function of a plane on the DSM image is defined as follows:

$$h = ax + by + c,$$

where $h$ is the height of the pixel on DSM, $(x, y)$ is the image coordinate of a pixel, and $a$, $b$, $c$ are parameters of plane function. For each size of the line segment, a set of $n$ pixels $S = \{(x_1, y_1, h_1), (x_2, y_2, h_2), \ldots, (x_n, y_n, h_n)\}$ are collected to fit for the plane function:

$$h_1 = ax_1 + by_1 + c$$
$$\vdots$$
$$h_n = ax_n + by_n + c$$

The corresponding residual equation is:

$$V = AX - L$$

with $V$ being the vector of least square residuals and $X$ being the plane parameters $\{a, b, c\}$, and

$$A = \begin{pmatrix} x_1 & y_1 & 1 \\ x_2 & y_2 & 1 \\ & \vdots & \\ x_n & y_n & 1 \end{pmatrix}, \qquad L = \begin{pmatrix} d_1 \\ d_2 \\ \vdots \\ d_n \end{pmatrix}$$

which can then be easily solved via least square algorithm. After the fitting of plane, we then calculate the adjusted DSM values for each pixel in the set $S$ by taking its position $(x, y)$ into the plane function. To make the DSM smooth after plane fitting, a feather operation is also implied. This adjustment procedure continues until all the line segments near the building boundary have been processed, which results the final adjusted DSM. Figure 4 shows the DSM before and after line based adjustment, we can see that the line based method can trim the original zigzag boundary well into straight line, which is much better in vision.

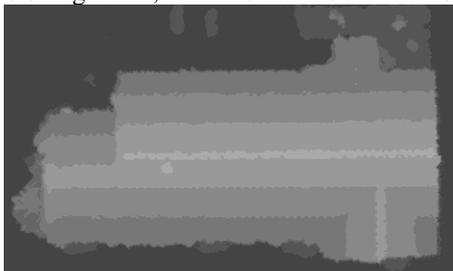 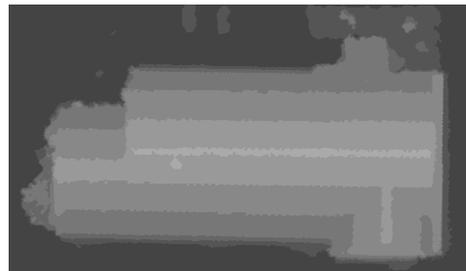

(a) original DSM  (b) DSM after adjustment

Figure 4 Comparison between the DSM before and after line based adjustment. The zigzag boundaries on the original DSM are adjusted well to straight line, the blur is due to the feather operation.

# EXPERIMENTS

**Test regions**

The proposed two methods have been tested on the ISPRS dataset on urban classification and 3D building reconstruction which contains DSM with a 9 cm ground sampling distance and corresponding three bands orthophoto. Also, the LiDAR point clouds in this region is provided, which makes it possible to perform quantitative analysis of the proposed methods. For testing, three patches with size 2048×2048 in pixels are cropped from the Vaihingen data (as Figure 5 shows), the corresponding area of each patch is thus 184m×184m. To test the robustness of the proposed methods, these patches are chosen in industrial region, sparse residential region and dense residual region, superlatively. For the industrial region, buildings are bigger and sparser than the residential region, which is easier for matching based DSM to generate good building boundary. While for the dense residual region, buildings are smaller and closer to each other, which makes it difficult to separate them in the matching based DSM.

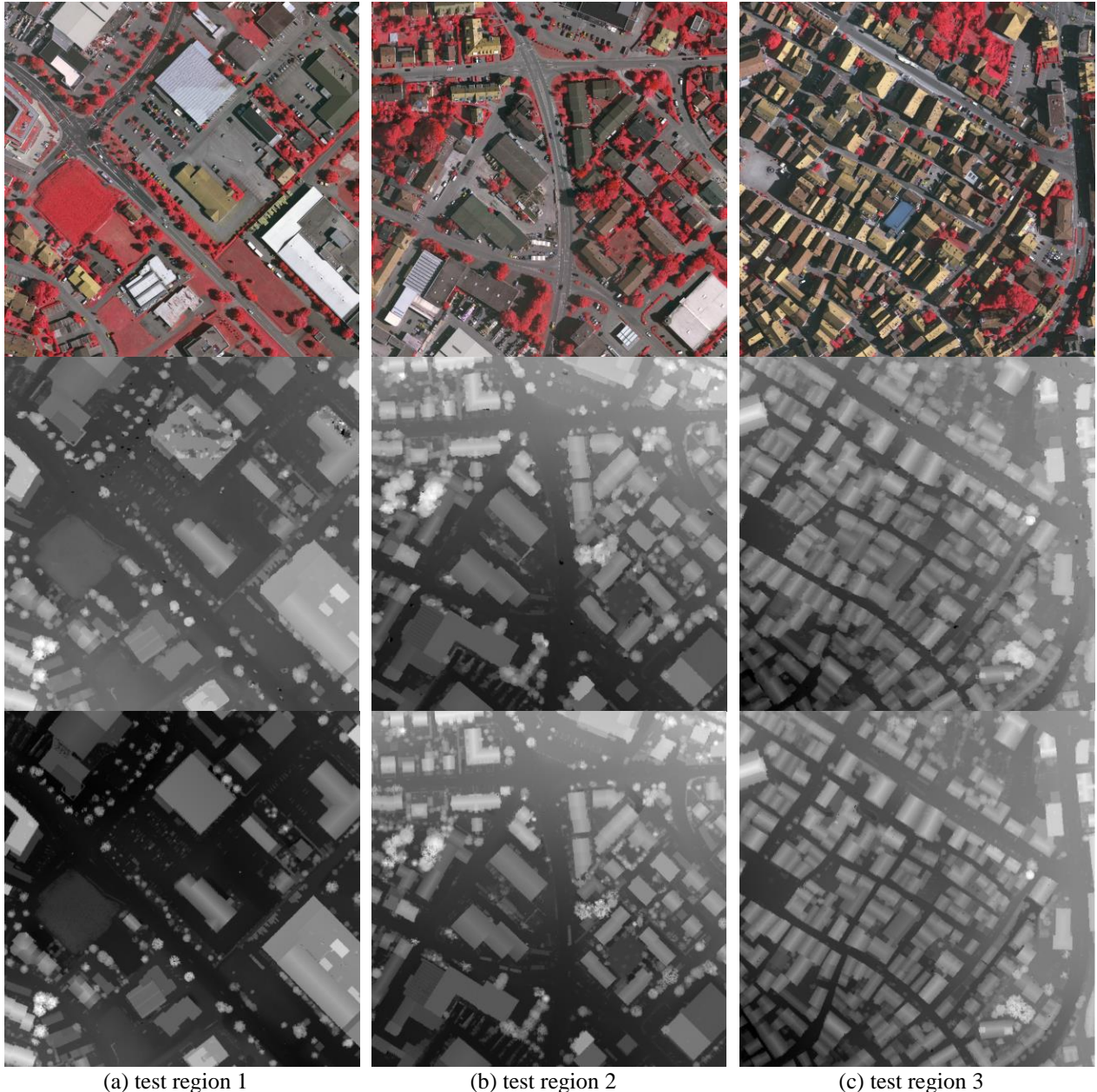

(a) test region 1    (b) test region 2    (c) test region 3

Figure 5 The testing regions utilized in our experiments. First row: the three bands color image (near infrared, red and green), second row: the corresponding DSM (9cm resolution) generated by matching, final row: the corresponding ground truth DSM (25cm resolution) generated from Lidar point clouds.

**Evaluation**

To analysis the performance of the methods quantitatively, we firstly generate a ground truth DSM from the given LiDAR point clouds. The resulting LiDAR-derived DSM thus has a ground sampling distance of 25 cm. Then given a computed DSM and the LiDAR-derived DSM, we calculate the RMSE of all the pixels on the DSM and those on the boundary, respectively. To choose pixels on the boundary, we generate a buffer zone for the boundaries with the buffer width equals to 5 pixels, 10 pixels, and 20 pixels, respectively to see the residual distribution pattern near the boundaries. Table 1 shows the RMSE errors between the DSM computer from original input (denoted as $DSM_{ori}$), DSM computed using line segment (denoted as $DSM_{line}$), and DSM computed via graph-cut (denoted as $DSM_{gc}$). Comparing each row, we can find that the RMSE increase from industrial region to dense residential region with the buildings getting smaller and closer. Comparing each column, we can see that the whole-image RMSE of $DSM_{line}$ is the smallest one among all the three DSMs. This is partially due to the reason that the plane fitting process has a averaging effect on the DSM values on the boundary, which may lead to small variance of DSM values and thus reduce the RMSE errors. The $DSM_{gc}$ gains the greatest whole-image RMSE, this is probably for the reason that the graph-cut method shifted more pixels than the other two methods (as can be seen in Figure 6). As to the RMSE of boundary pixels, both methods proposed in this work achieves better result than the original DSM, with an averaging improvement of 2.3% and 3.2%, respectively. The $DSM_{gc}$ happens to gain the smallest RMSE in all the tests. A reasonable explanation can be that the smooth constrains in graph-cut makes the pixels on each contour are adjusted as a whole, thus more boundary pixels will be modified to shift toward the boundary than the line segment based method which refine each line segment individually. Figure 6 is a demonstration of the relationship between the RMSE and the buffer size of the building boundary. Combining Figure 6 and Table 1, we can see that on the boundary region, the RMSE are constantly reduced with the increasing of the buffer size, which is consistent with the previous observation that the RMSE on the boundary is greater than that on the whole region.

Figure 7 shows the overall and detailed RMSE error distribution of $DSM_{ori}$, $DSM_{line}$, and $DSM_{gc}$ on the second testing region, respectively. As we can see from the second row that both the proposed method can reduce the RMSE on the boundary. The difference is that, the line segment based methods tends to flatten the boundary region to reduce the RMSE because of its plane fitting procedure, while the graph-cut based method tends to generate uneven boundaries due to its pixelwise optimization procedure.

To further demonstrate the effect of the proposed DSM boundary adjustment methods, Figure 8 shows the detailed pixel-level cross-section analysis of two boundaries. As can be found in Figure 8, the both the two proposed methods have significantly improved the RMSE. For the first case, the RMSE is improved c

Table 1. RMSE errors between the computed DSM and the LiDAR-derived DSM (meter)

| Test Region | $DSM_{ori}$ | | | | $DSM_{line}$ | | | | $DSM_{gc}$ | | | |
|---|---|---|---|---|---|---|---|---|---|---|---|---|
| | Whole Image | # 5 pixels | # 10 pixels | # 20 pixels | Whole Image | # 5 pixels | # 10 pixels | # 20 pixels | Whole Image | # 5 pixels | # 10 pixels | # 20 pixels |
| Region 1 | 1.246 | 1.853 | 1.794 | 1.702 | 1.236 | 1.789 | 1.744 | 1.667 | 1.258 | 1.791 | 1.734 | 1.667 |
| Region 2 | 1.460 | 1.849 | 1.772 | 1.636 | 1.446 | 1.771 | 1.710 | 1.590 | 1.487 | 1.733 | 1.679 | 1.577 |
| Region 3 | 2.140 | 2.609 | 2.473 | 2.324 | 2.134 | 2.579 | 2.458 | 2.313 | 2.146 | 2.561 | 2.432 | 2.298 |

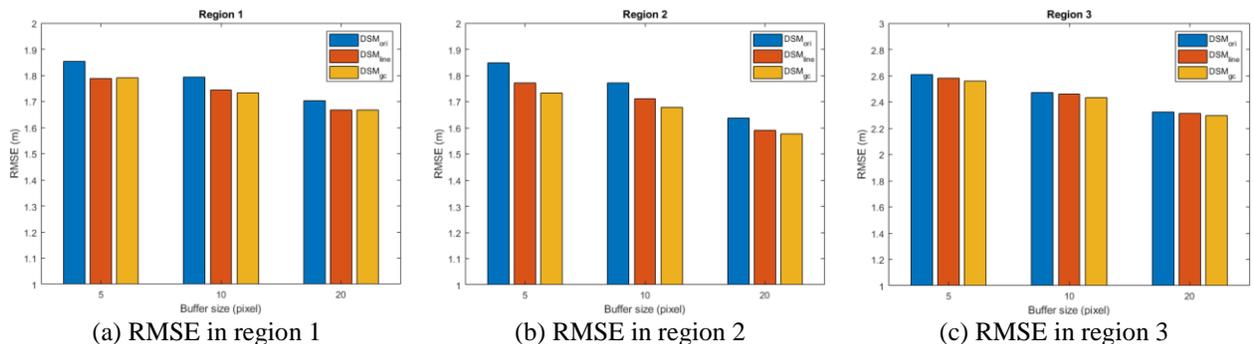

(a) RMSE in region 1      (b) RMSE in region 2      (c) RMSE in region 3

Figure 6 Variation of RMSE with the buffer size of building boundary on three testing regions. It can be seen that the RMSE reduces with the increasing of the buffer size.

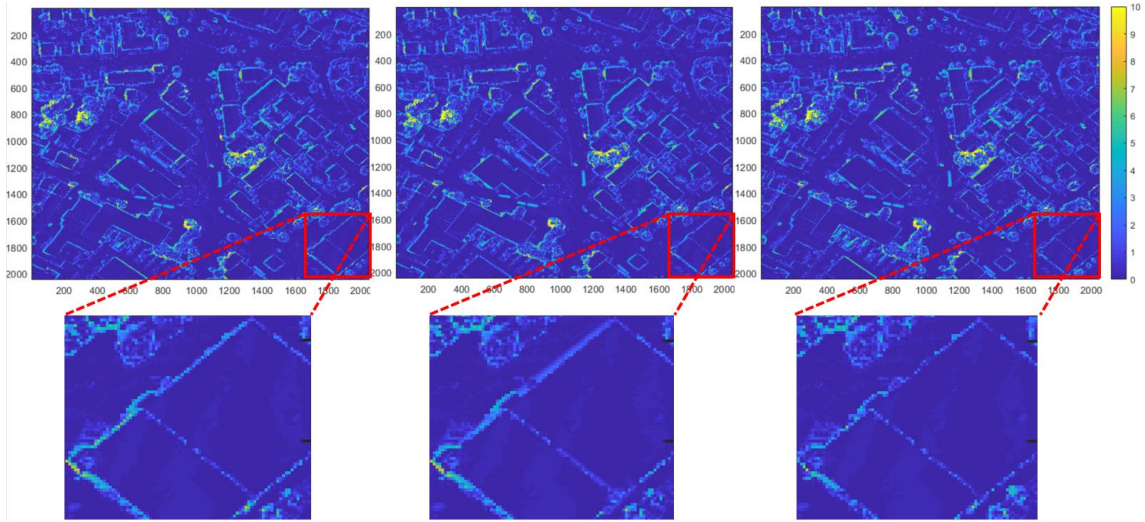

(a) RMSE of $DSM_{ori}$      (b) RMSE of $DSM_{line}$      (c) RMSE of $DSM_{gc}$

Figure 7 The RMSE error of the original DSM, the DSM refined by line segment, and the DSM refined via graph-cut on test region 2. First row: the RMSE error on the whole image. Second row: details of RMSE error on a building.

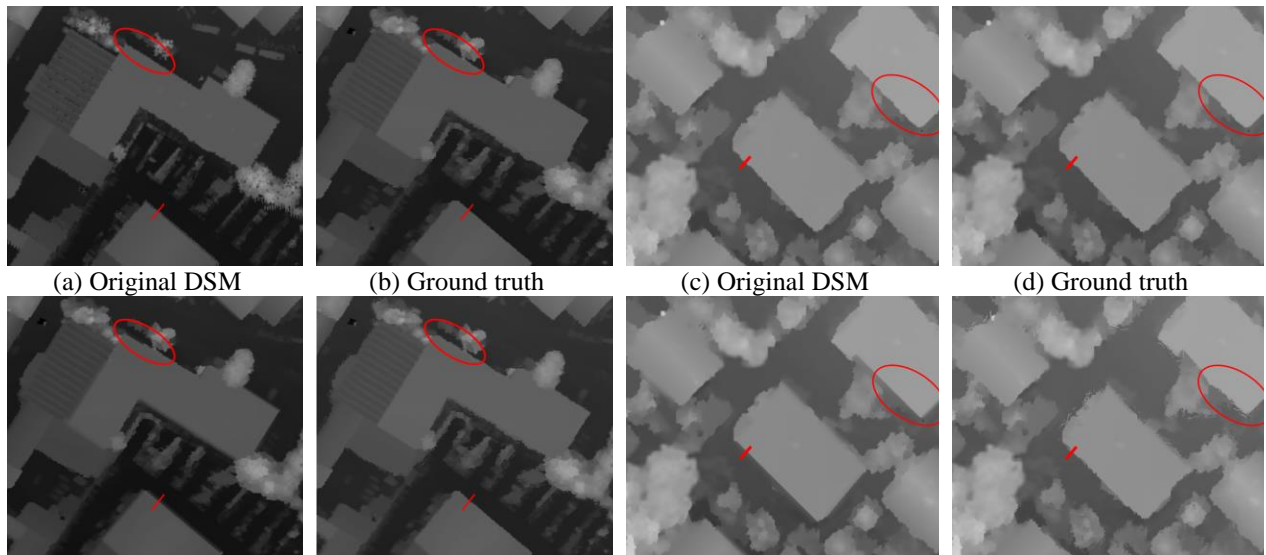

(a) Original DSM    (b) Ground truth    (c) Original DSM    (d) Ground truth

(e) Line based DSM    (f) Graph-cut DSM    (g) Line based DSM    (h) Graph-cut DSM

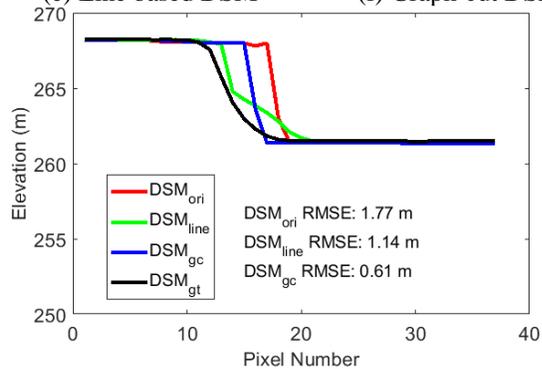
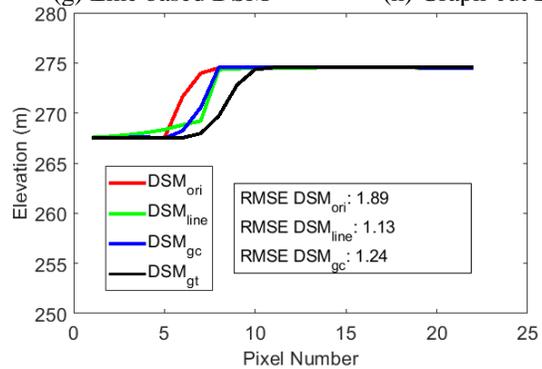

(i) Cross-section analysis      (j) Cross-section analysis

Figure 8 Cross-section analysis of two building boundaries. The left two columns are results of the first boundary, the right two columns belong to the second one. (i) and (j) demonstrate the elevation of pixels on different DSMs and the corresponding RMSE values for two testing boundaries, respectively.

## 4. CONCLUSION

In this paper, we propose two methods to utilize the line segments extracted from the orthophoto to adjust the corresponding DSM. One of the methods uses graph-cut to minimize the distance between the DSM boundaries to the line segments. The other method fits plane on both sides of the line segment to rectify the pixel values on the DSM. Both methods have been tested on the ISPRS dataset on urban classification and 3D building reconstruction. Experimental results show that both methods can improve the RMSE of the DSM, especially on the building boundaries where the RMSE is promoted from 1.77m to 1.14m and 0.61m for one case, from 1.89m to 1.13m and 1.24m for the other case. However, we are also aware that the proposed method is a local method and the results are largely dependent on the extracted building boundaries and line segments. Therefore, in the future work we will focus on more robust boundary and line segment detectors, for example, the deep learning based methods.

**Disclaimer:** Mention of brand names in this paper does not constitute an endorsement by the authors.

## ACKNOWLEDGEMENT

The study is partially supported by the ONR grant (Award No. N000141712928). We would like to thank the German Society for Photogrammetry, Remote Sensing and Geoinformation (DGPF) for providing the Vaihingen data set, and the Optech Inc., First Base Solutions Inc., York University, and ISPRS WG III/4 for providing the Downtown Toronto data set.

## REFERENCES

Bobick, Aaron F., and Stephen S. Intille. "Large occlusion stereo." International Journal of Computer Vision 33.3 (1999): 181-200.

Geiger, Andreas, Martin Roser, and Raquel Urtasun. "Efficient large-scale stereo matching." Asian Conference on Computer Vision (ACCV), 2010.

Brédif, Mathieu, et al. "Extracting polygonal building footprints from digital surface models: a fully-automatic global optimization framework." ISPRS journal of photogrammetry and remote sensing 77 (2013): 57-65.

Lu, Xiaohu, et al. "Cannylines: A parameter-free line segment detector." IEEE International Conference on Image Processing (ICIP), 2015.

Von Gioi, Rafael Grompone, et al. "LSD: A fast line segment detector with a false detection control." IEEE transactions on pattern analysis and machine intelligence 32.4 (2010): 722-732.

Boykov, Yuri, Olga Veksler, and Ramin Zabih. "Fast approximate energy minimization via graph cuts." IEEE Transactions on pattern analysis and machine intelligence 23.11 (2001): 1222-1239.

Hirschmuller, Heiko. "Stereo processing by semiglobal matching and mutual information." IEEE Transactions on pattern analysis and machine intelligence 30.2 (2008): 328-341.

Huang, Xiaoming, and Yu-Jin Zhang. "An O (1) disparity refinement method for stereo matching." Pattern Recognition 55 (2016): 198-206.

Park, Se-Hoon, Min-Gyu Park, and Kuk-Jin Yoon. "Confidence-based weighted median filter for effective disparity map refinement." IEEE International Conference on Ubiquitous Robots and Ambient Intelligence (URAI), 2015.

Li, Hui, Xiao-Guang Zhang, and Zheng Sun. "A line-based adaptive-weight matching algorithm using loopy belief propagation." Mathematical Problems in Engineering 3(2015):1-13.

Qin, Rongjun, et al. "Disparity Refinement in Depth Discontinuity Using Robustly Matched Straight Lines for Digital Surface Model Generation." IEEE Journal of Selected Topics in Applied Earth Observations and Remote Sensing (2018).